\documentclass{article}

     \usepackage[nonatbib,preprint]{neurips_2021}

\usepackage[utf8]{inputenc} 
\usepackage[T1]{fontenc}    
\usepackage{hyperref}       
\usepackage{url}            
\usepackage{booktabs}       
\usepackage{amsfonts}       
\usepackage{nicefrac}       
\usepackage{microtype}      
\usepackage{xcolor}         

\usepackage[pdftex]{graphicx}
\usepackage{wrapfig}
\usepackage{algorithm}
\usepackage[noend]{algpseudocode}
\usepackage{enumitem}
\usepackage{caption}
\usepackage{subcaption}
\usepackage{multirow, makecell}
\usepackage{amsmath}
\usepackage[noadjust]{cite}

\title{Search Spaces for Neural Model Training}

\author{%
  Darko Stosic \\
  NVIDIA\\
  \texttt{darkos@nvidia.com} \\
  \And
  Dusan Stosic \\
  NVIDIA \\
  \texttt{dstosic@nvidia.com} \\
}

\begin{document}

\maketitle

\begin{abstract}
While larger neural models are pushing the boundaries of what deep learning can do, often more weights are needed to train models rather than to run inference for tasks. This paper seeks to understand this behavior using search spaces -- adding weights creates extra degrees of freedom that form new paths for optimization (or wider search spaces) rendering neural model training more effective. We then show how we can augment search spaces to train sparse models attaining competitive scores across dozens of deep learning workloads. They are also are tolerant of structures targeting current hardware, opening avenues for training and inference acceleration. Our work encourages research to explore beyond massive neural models being used today.
\end{abstract}

\section{Introduction}
In the area of deep learning, increasing the size of neural models has led to dramatic advances~\cite{hestness2017,kaplan2020,henighan2020}, motivating training of hundreds of billions of parameters~\cite{brown2020,fedus2021}. However, the growing size of models incur higher memory costs and runtimes, hindering training or inference tasks that can be tackled using current hardware~\cite{thompson2020}.
Nevertheless, larger models have been shown to train faster~\cite{li2020} and better~\cite{kaplan2020} which drives their adoption.

An area of research that seeks to compensate the increasing costs of model size is sparsity~\cite{sparsitysurvey}, which moves weights to zero so they can be discarded from storage or computations. Interestingly, neural models are capable of performing tasks on sizes smaller than they were trained for~\cite{narang2017,renda2020}. While sparsity emerges as a promising option to reduce inference costs for overparameterized neural models~\cite{thinet,allenzhu2019}, training observes limited success~\cite{snip,lottery,grasp,syncflow,frankle2020,bellec2018,mocanu2018,dettmers2019,rigl,topkast}. This raises an interesting question -- why do larger neural models train better -- that remains to be understood. Being able to run inference but not training using fewer weights points to a problem with search, or how models navigate search spaces throughout training, rather than model capacity~\cite{lecun1990}.

This paper explains why deep learning tasks improve with the size of search spaces being explored during training. We find that adding weights provides extra degrees of freedom that form new paths of optimization and facilitate the search for neural models. Using this understanding, we propose a series of steps to augment search spaces of sparse models during training to approximate the behavior of larger models\footnote{We will make the code available in the near future.}. We then show this methodology achieves competitive results on dozens of deep learning workloads, even when satisfying constraints needed to accelerate training and inference using Sparse Tensor Cores~\cite{mishra2021} in NVIDIA GPUs.

The paper is organized as follows. In Section~\ref{sec:searchspace} we use search spaces to describe reasons more weights are needed for training. We then devise approximations to augment their size when training sparse models in Section~\ref{sec:considerations}. Section~\ref{sec:experiments} shows sparse models constructed in this fashion perform competitively across a plethora of deep learning workloads and can target current hardware accelerators~\cite{mishra2021} for training and inference. In Section~\ref{sec:relatedwork} we relate to prior research. Section~\ref{sec:conclusion} concludes with directions for future work.

\section{A primer on search spaces}\label{sec:searchspace}
We begin by understanding what role search spaces (or the set of all possible weight combinations that a model can take) have in neural model training. We show neural weights needed for inference are discovered throughout training with help from added weights (or weights that can be discarded after training). More specifically, we observe adding weights help neural models better explore search spaces to discover promising combinations for inference. While they do not learn meaningful representations, added weights form alternate paths for optimization that enable training to escape critical points. By combining these observations, we provide intuition that more weights are needed for training because they augment search spaces allowing neural models to train more effectively.

\begin{figure}[!t]
\begin{center}
\includegraphics[width=\linewidth]{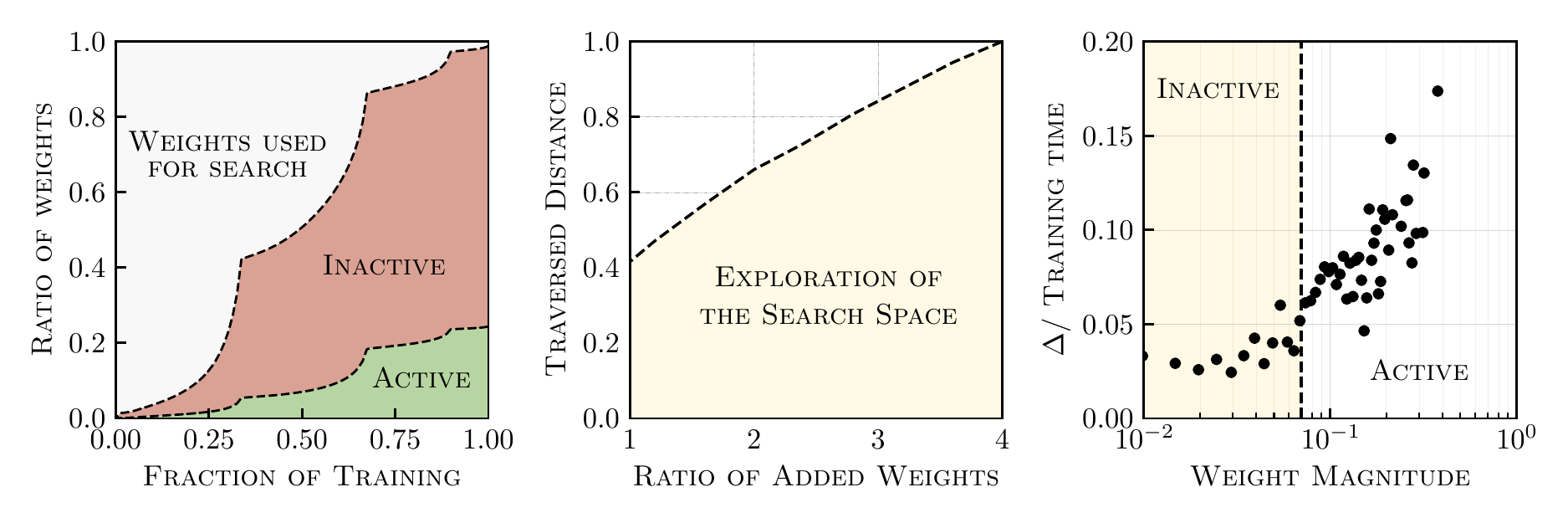}
\caption{Investigations into the role of search spaces for ResNet50. Left: Fraction of weights whose magnitudes remains above (active set) or below (inactive set) a threshold until the end of training. The threshold is chosen such that one-fourth of the weights remain for inference tasks.
Middle: Cumulative distance (normalized by its maximum) that neural models traverse in search space as a function of the number of weights added to the active set. Right: The time ($\Delta$) it takes for weights to uncorrelate (normalized by the total number of training steps) as a function of magnitudes obtained after training. Points correspond to the median $\Delta$ over weight bins sampled from a single neural layer. Appendix~\ref{sec:suppsearchspace} covers data for a broader span of neural models and tasks.
}
\label{figsearch}
\end{center}
\end{figure}

\subsection{Discovering weights needed for inference}\label{sec:discovering}
A popular technique to reduce inference costs is to discard weights of a neural model whose magnitudes fall below a certain threshold after training (typically denoted as sparsity), since they contribute the most to defining the behavior of a neural model~\cite{topkast}. Neural models constructed in this fashion (or sparse models) can achieve the same accuracy as larger models when trained from the same initialization~\cite{lottery} or with additional training steps~\cite{mishra2021}. Thus, we can assume neural weights retained after training are critical for inference, as opposed to removed weights which are not necessary.

The question then arises whether weights relevant for inference can be determined earlier in the training process. For this purpose, we examine how weights evolve throughout training, classifying them into two types of behavior. We define a set of weights as active at a particular point in time if their magnitudes remain above a chosen threshold from that point until the end of training, representing weights that are important for inference. Conversely, the set of weights whose magnitudes remain below the threshold from that point until the end of training are inactive, after which they play no role in inference tasks. There are also the remaining weights that constantly move above and below the threshold. While other combinations of weights that can contribute to inference may exist, this is one instance that is known to produce good accuracy~\cite{lottery}.

Figure 1 (left) shows how sets evolve over the course of training when using a threshold that maintains a quarter of the weights for inference. In early training, most weights switch between either side of the threshold as neural models are still learning, eventually stabilizing into active or inactive sets. Neural weights relevant for inference (or form the active set) are not prebaked~\cite{frankle2020}, but rather discovered as the model learns. As a result, we can presume all weights including ones that are eventually discarded (or added weights) play a crucial role in determining what will be used for inference.

\subsection{Exploring search spaces with model size}
An important aspect of training is how well can neural models navigate search spaces to discover better weight combinations for inference tasks. Given added weights help uncover weights relevant for inference, we posit they must play an important role in exploring search spaces. We characterize exploration with distances that neural models undertake throughout training, which are measured using cumulative increments $\sum_t\sum_w|w(t+1)-w(t)|$ over successive time steps $t$ across convolutions and linear layers, excluding other neural layers.

Figure~\ref{figsearch} (middle) illustrates the distance traversed (or exploration of the search space) as a function of weights added to a model that can successfully perform inference tasks (assume a model one-fourth its original size) but requires more weights to train. We observe distance increases with the number of weights being added, therefore larger models can navigate search spaces more effectively.

Based on the above, we reason using more weights during training expands the size of search spaces, making it easier to discover combinations of weights that are needed for inference. More precisely, adding weights helps neural models explore new regions of the search space, escaping critical points that often lead to poor task accuracy in smaller models~\cite{choromanska2015}. However, the exact function of added weights (which can be later discarded for inference) is discussed below.

\subsection{Roles of weights using similarities} \label{correlations}
To better understand the role of weights during training, we look at how similar are their values over time using correlations. We can interpret correlations as the degree with which weights in a neural model learn: weights that get reinforced (or learn) throughout training are correlated to previous values and exhibit temporal patterns, and conversely weights that have random changes (or don't learn) are uncorrelated and have no dependence over time.

We measure the similarity between a pair of series $x$ and $y$ through the Pearson correlation coefficient~\cite{pearson}: $\rho_{x,y}\equiv E[(x-\mu_x)(y-\mu_y)]/\sigma_x\sigma_y$, where $\mu$ is the mean, $\sigma$ the standard deviation, and $E[\cdot]$  the expected value. Coefficients of value  $+1$ denote two series have identical trends, $0$ indicates the series are random, and $-1$ represents series with opposite behavior. We treat each individual weight as a time series $w\in\{w_1,w_2,\dots,w_t\}$ and compute Pearson coefficients between windows $x\in\{w_1,w_2,\dots,w_{t-\tau}\}$ and $y\in\{w_{\tau},w_{\tau+1},\dots,w_{t}\}$, representing the similarity between the current weights and their values at some future time $\tau$. Since correlations naturally decay with $\tau$ (temporal patterns are less likely to persist over long durations), an important metric is the time $\Delta$ after which weights no longer correlate to their future values ($\rho=0$).

Figure~\ref{figsearch} (right) illustrates the fraction of total training after which weights become completely uncorrelated ($\Delta$) as a function of their magnitudes obtained after training. While weights belonging to the active set exhibit correlations that persist across a significant portion of training, weights forming the inactive set have short-term correlations and behave randomly over small periods. We conjecture long-term correlations of large weights are characteristic of learning, as they signify repeated reinforcement along the weight direction, whereas the near-random motion of small weights suggests they don't learn useful representations and can be removed during inference.

Since weights that move into the inactive set are absent of correlations across long periods of training, we interpret their crucial role during training as follows: short-term interactions enable training to take different paths for optimization, which facilitates escape from bad saddle points~\cite{kawaguchi2016} or high error plateaus~\cite{dauphin2014} that can slow down learning. In later sections, we also discover their learned values can be periodically destroyed throughout training without affecting accuracy. Thus, we reasonably conclude that added weights removed during inference do not learn meaningful representations, but rather their short-term behavior is responsible for widening the search space.

\subsection{An intuitive explanation about search spaces}
Based on the data above, we construct the hypothesis that ``weights removable at inference time represent extra degrees of freedom that are needed to augment search spaces for training neural models effectively.'' This explains the prevailing observation that neural models with reduced sizes, which perform well at inference tasks, would underperform during training.

\begin{wrapfigure}[17]{r}{0.5\textwidth}
\vspace{-15pt}
\begin{center}
\includegraphics[trim=140 50 50 55,clip,width=0.45\textwidth]{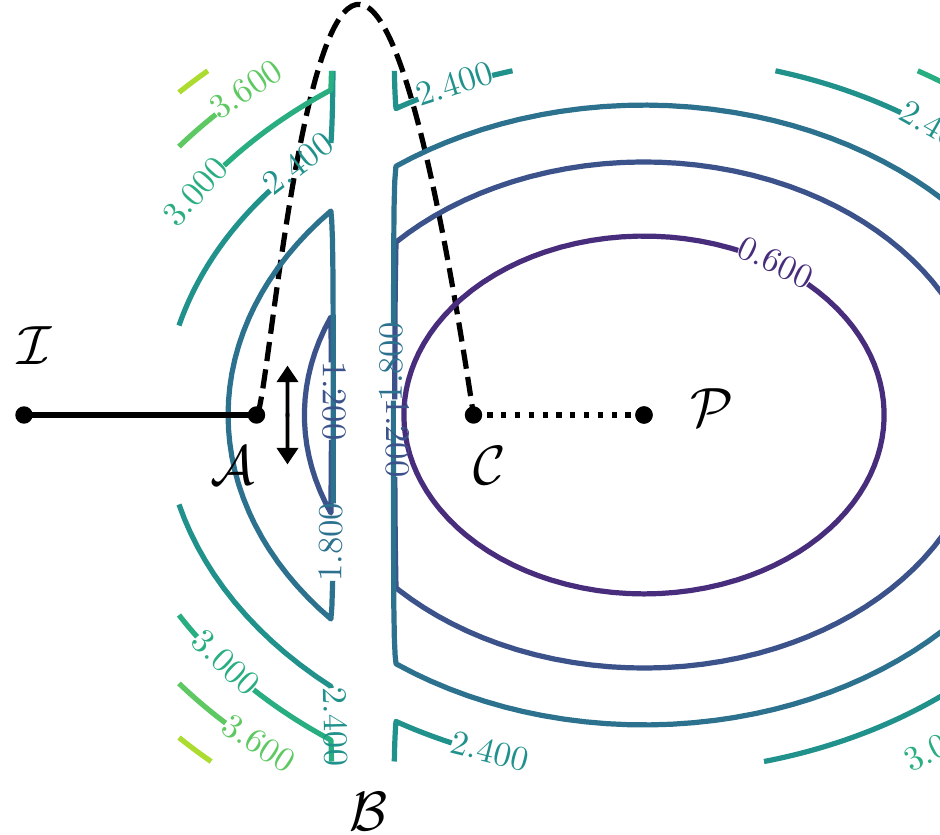}
\caption{Training trajectories for one- and two-parameter neural models on a 2D convex landscape with a minimum $\mathcal{P}$ and barrier $\mathcal{B}$.}
\label{figsketch}
\end{center}
\end{wrapfigure}
For a more precise formulation, assume neural models are parameterized by $n$. In this scenario, training traverses through $n$-dimensional search spaces to minimize the loss function, leveraging all $n$ weights in the process. Conversely, models trained with $dn$ weights, where $d\in(0,1)$, navigate search spaces that are constrained to $dn$ dimensions. Operating on reduced spaces makes training more susceptible to critical points, limiting weight combinations that neural models can explore during training. Since weights removable at inference observe short-term interactions, they do not learn representations but rather augment search spaces allowing training to be more effective.

Figure~\ref{figsketch} illustrates an example. Let's assume we are training neural models in a two-dimensional loss landscape of convex shape with a minimum at $\mathcal{P}$ and a vertical barrier $\mathcal{B}$. A one-parameter model that can move left or right during training will converge prematurely at $\mathcal{A}$ if initialized at $\mathcal{I}$ such that $\mathcal{B}$ obstructs its path to $\mathcal{P}$ (solid curve). Adding another weight augments the search space so training can take new possible paths along the loss landscape to escape $\mathcal{B}$ (dashed curve). After reaching $\mathcal{C}$, training can discard the added weight and converge on $\mathcal{P}$ (dotted curve). A similar generalization can be made for higher dimensions, where adding weights helps training escape saddle points. This behavior is believed to be a crucial aspect for neural model training with backpropagation.

\section{Recommendations for sparse training}\label{sec:considerations}
While search spaces are important for training, their sizes are often made smaller for sparse models. Building on understanding from previous sections, we uncover the following recommendations to approximate wider search spaces when training sparse models:
\begin{enumerate}[label=(\arabic*)]
    \item Rewiring weights that participate in training allows sparse models to explore different regions of the search space.
    \item Gradient updates for weights that do not participate in training encourages alternate paths for optimization.
    \item Inducing exploitation (stop rewiring after some amount of training steps, reset non-participating weights to zero or regularize them) reduces noise from gradient accumulations.
\end{enumerate}
Appendix~\ref{sec:methods} summarizes possible methods for expanding the size of search spaces during neural model training using these three simple steps.

\begin{figure}[!t]
\begin{center}
\includegraphics[width=\linewidth]{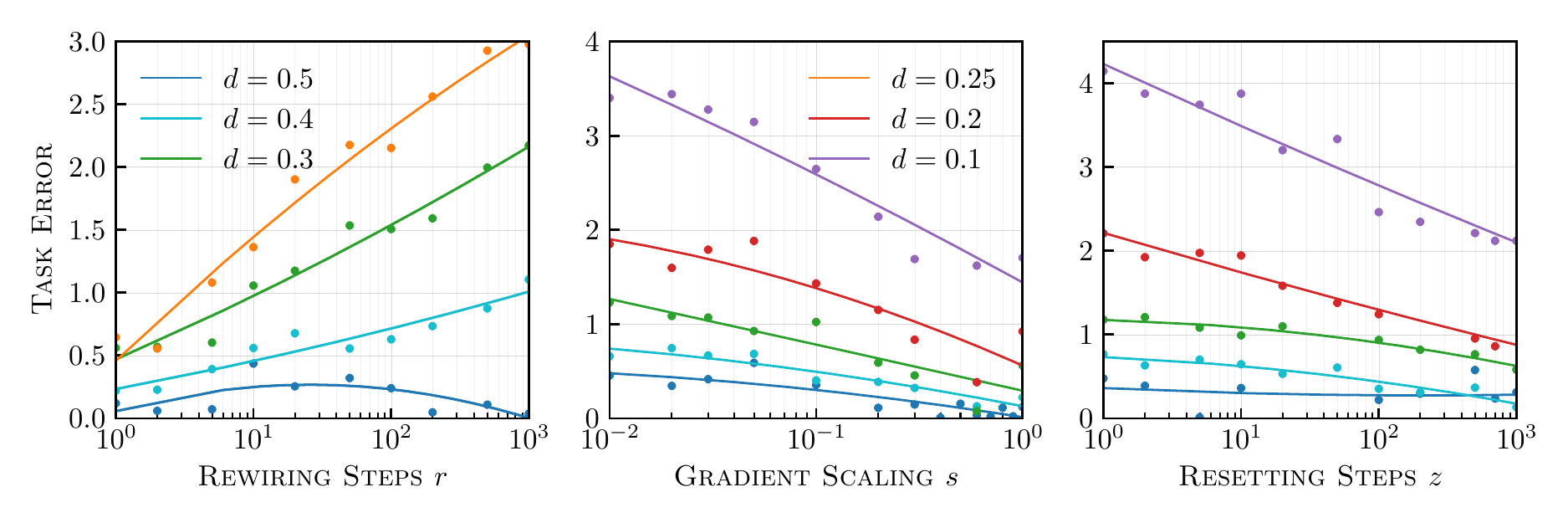}
\caption{Investigations into various aspects of training sparse models using ResNet50. Left: Task error (or accuracy difference between regular and sparse models) as a function of rewiring steps $r$. Middle: Task error as a function of scaling factor $s$ applied to gradients of weights that do not participate in training. Right: Task error where non-participating weights are reset to zero every $z$ training steps. We consider sparse models of different sizes $d$ denoting the ratio of weights being used. Lines represent polynomial fits of sample points. Appendix~\ref{sec:suppenhance} covers data for a broader span of neural models and tasks.
}
\label{figenhance}
\end{center}
\end{figure}

\subsection{Rewiring of neural weights}
Sparse models are commonly trained by rewiring (or sampling) a subset of weights from a neural model according to some criteria (magnitude, sign, gradients, etc), allowing them to learn which weights are needed for inference. Some works rewire weights every training iteration~\cite{bellec2018,neuralwirings,swat,topkast,nmsparsity}, while others rewire every hundreds of training steps~\cite{rigl} or after an entire pass through the data~\cite{mocanu2018,dettmers2019}. Since there are no clear reasons behind these choices, an important question becomes how often must weights be rewired so neural models can learn.

We study the effects rewiring rates have on accuracy when training sparse models of various sizes $d$, where $d$ denotes the fraction of weights being used. Figure~\ref{figenhance} (left) plots the task error (or accuracy difference between regular and sparse models) as a function of the number of training steps $r$ taken between rewirings. We find rewiring frequency does not matter when sparsity is low or moderate ($d\sim0.5$), but errors increase with $r$ as models get sparser ($d\ll0.5$). Therefore, weights should be rewired frequently (within a few training steps) to avoid losing accuracy.

We can use search spaces to explain why training improves with more frequent rewiring. A sparse model is made of weights that participate in training (are used in forward and backward propagations) and those that do not participate (are set to zero during training). Search spaces will expand only when weights are rewired: neural models always operate on reduced spaces if participating weights remain unchanged throughout training ($r\rightarrow\infty$), and expand exactly once after going through the entire data when they are rewired every epoch. By swapping between weights that participate and do not participate, training can take different paths for optimization and approximate a free search space, which helps neural models escape critical points~\cite{evci2019}.

\subsection{Updates for weights not participating in training}\label{sec:gradientupdates}
Previously, we found neural weights not needed for inference form short-term interactions that make training more effective. Since sparse models lack weights that have similar roles, we can approximate this behavior using gradient updates for non-participating weights. Because non-participating weights do not contribute to the loss, their gradients determine whether another optimization path is better suited to traverse the region being explored in the loss landscape. Repeating gradient updates registers the importance of these paths over some period of training and can trigger rewiring when weights exceed a threshold, which allows training to explore different regions of the search space while operating on (or forward and backward propagating) a smaller set of weights.

We evaluate the importance of gradients updates on non-participating weights by reducing their contribution with a scale factor $s$. Figure~\ref{figenhance} (middle) shows the task error as a function of the scale factor $s$ using various sizes $d$. We observe error increases as $s\rightarrow0$ (no gradients contribute when $s=0$), which can be attributed to premature convergence when training lacks expressive power to explore different paths. Dampened gradient updates affect sparser models ($d\ll0.5$), which are sensitive to reduced spaces, more than larger models ($d\sim0.5$), which may still have sufficient search. As a result, non-participating weights should be updated regularly to retain accuracy.

Gradient updates for weights that do not participate in training could mean they also learn representations, such that more capacity rather than search improves accuracy. We verify this by resetting non-participating weights to zero after every $z$ training steps, thus removing any representations they might have learned. Figure~\ref{figenhance} (right) shows the task error as a function of $z$. Errors are the highest when values are reset every iteration ($z=1$), which is equivalent to training on reduced spaces (weights can still be rewired but we no longer encourage certain ones to participate in training). Conversely, error decreases as weights are reset less frequently and saturates after sufficient training steps ($z\sim 1k$). Interestingly, this also represents the time it takes for added weights to uncorrelate, as shown in previous sections. The ability to remove information from non-participating weights reinforces the idea that they augment search spaces rather than model capacity.

While most of literature trains sparse models using fewer weights and gradients~\cite{bellec2018,mocanu2018,dettmers2019}, more recent success has been found by updates gradients for weights that do not participate in training~\cite{neuralwirings,dst,nmsparsity,hubara2021} as described above.

\subsection{Balancing between exploration and exploitation}\label{sec:exploitation}
Deep learning like other optimization problems treads a delicate balance between exploration and exploitation. In early stages of training neural models explore search spaces using high learning rates, whereas late stages exploit specific regions using small learning rates. Training sparse neural models can also affect this balance, as fewer weights reduces the degrees of freedom and thus hinders exploration during training. On the other hand, gradient updates on non-participating weights introduces noise as any nonzero value will not reflect what is being used for training, which limits exploitation (training bounces around basins of attraction due to gradient noise).
\begin{wrapfigure}[20]{r}{0.5\textwidth}
\begin{center}
\includegraphics[trim=10 10 10 10,clip,width=0.44\textwidth]{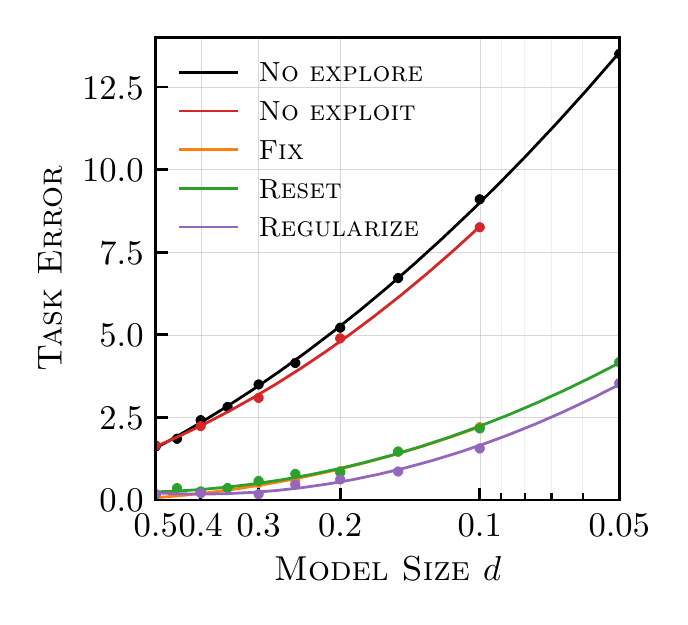}
\caption{Task error after training ResNet50 using different exploration versus exploitation strategies as a function of model size $d$. Appendix~\ref{sec:suppenhance} covers data for a broader span of neural models and tasks.}
\label{figexplorationresnet}
\end{center}
\end{wrapfigure}

Figure~\ref{figexplorationresnet} shows task error degrades without proper exploration, by training smaller models with neural layers of reduced widths (\textproc{No explore}), or exploitation, by training sparse models delineated so far (\textproc{No exploit}). Therefore, we seek ways to induce exploitation when augmenting search spaces for more exploration. One course of action is to remove noise introduced by non-participating weights during late stages, so training can take steepest descents towards the minima. To achieve this we stop rewiring weights after sufficient training (\textproc{Fix}), such that non-participating weights can no longer contribute to the loss. Figure~\ref{figexplorationresnet} shows this decreases error rates tremendously.

Another option is to draw analogies with added weights, which can be safely discarded after sufficient training since they do not learn representations over time. We reset non-participating weights to zero roughly every $1k$ training steps (or the time it takes for added weights to uncorrelate) in order to remove gradient noise that may trigger unnecessary rewiring, allowing training to exploit nearby regions. Figure~\ref{figexplorationresnet} shows error decreases substantially when non-participating weights are either reset to zero (\textproc{Reset}) or regularized with a decay factor (\textproc{Regularize}) as suggested in~\cite{nmsparsity}. Sparsity literature introduces alternative courses for inducing exploitation~\cite{mocanu2018,dettmers2019,rigl}.

\section{Empirical data}\label{sec:experiments}
Using recommendations from the previous section, we are in position to train sparse models approximating wider search spaces. This section presents empirical evidence that wider search spaces can help sparse models achieve better accuracy for inference tasks. First, we demonstrate that our strategy performs competitively against state-of-the-art sparsity research. Then, we explore the effects of sparsity on models that are trained to the limits of their capacity. Lastly, we discover sparse models are tolerant to sparsity structures that target hardware acceleration using Sparse Tensor Cores~\cite{mishra2021}.

\subsection{Comparisons to sparsity research}
Experiments are conducted across a wide breadth of deep learning tasks and neural architectures trained on large data sets (Appendix~\ref{sec:setup} details the setup). We draw comparisons between full-sized models that have free search (\textproc{Regular}), smaller models consisting of neural layers with reduced widths (\textproc{Reduce}), and sparse models that approximate wider search spaces (\textproc{Search}). Appendix~\ref{sec:capacity} shows where accuracy for the latter falls between the former two.

Table~\ref{tabunstruct} lists accuracies for regular models and their differences for sparse models (adding the two numbers produces accuracies for sparse models) across various tasks as a function of $d$, where $d$ denotes the ratio of weights being used. We find using our recommendations most sparse models can be halved in size ($d=0.5$) without sacrificing any task accuracy, whereas training with a quarter of the weights ($d=0.25$) often reduces accuracy by less than $1\%$. Some exceptions include efficient convolutional models and sparser models ($d=0.1$) which may be constrained by capacity.

\begin{figure}[!t]
\begin{center}
\includegraphics[width=\linewidth]{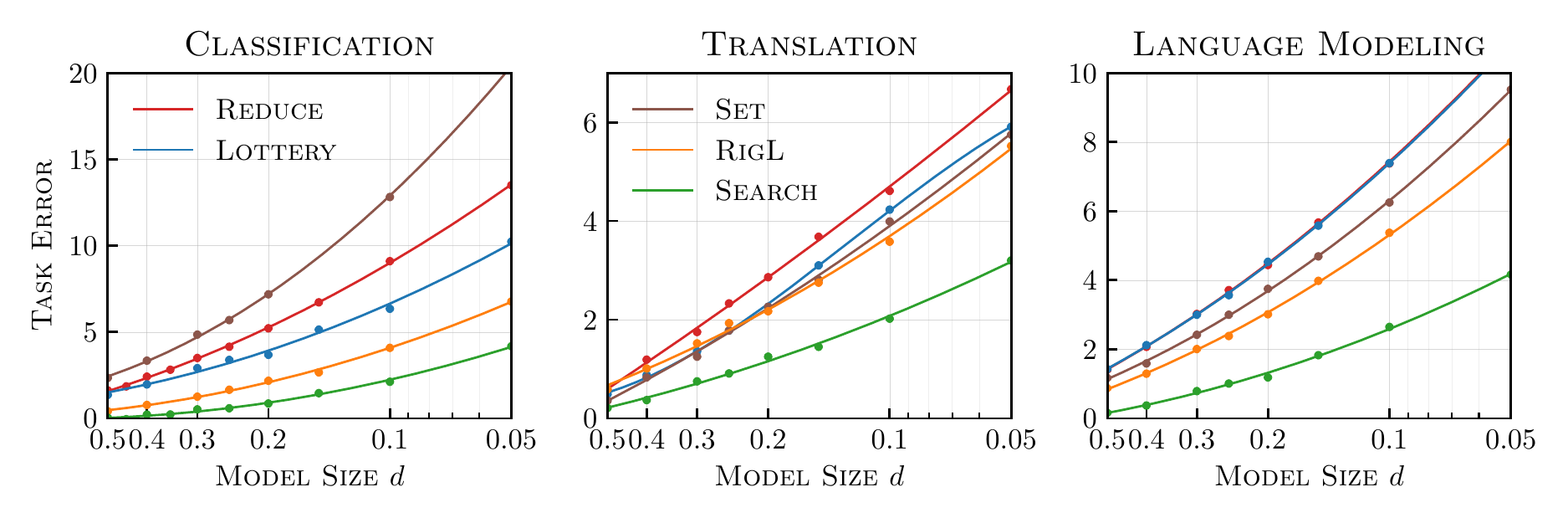}
\caption{Task error (or accuracy difference between regular and sparse models) comparing various methods as a function of model size $d$. Left to right: ResNet50 (Classification), Transformer (Translation), Transformer-XL (Language Modeling). Appendix~\ref{sec:suppmethods} covers data for a broader span.
}
\label{figmethods}
\end{center}
\end{figure}

\begin{table}[!b]
\caption{Accuracies for regular models and their differences for sparse models using different sizes $d$.}
\label{tabunstruct}
\begin{center}
\resizebox{\textwidth}{!}{%
\begin{tabular}{lcccclcccc}
\toprule
Model & \textproc{Regular} & $d=0.5$ & $d=0.25$ & $d=0.1$ & Model & \textproc{Regular} & $d=0.5$ & $d=0.25$ & $d=0.1$ \\
\midrule
ResNet18	&	$70.30$	&	$+0.01$	&	$-0.92$	&	$-2.75$	&	SqueezeNet V1	&	$60.77$	&	$-0.88$	&	$-5.20$	&	$-$	\\
ResNet34	&	$73.87$	&	$-0.20$	&	$-0.65$	&	$-2.27$	&	MobileNet V2	&	$71.53$	&	$-0.87$	&	$-0.87$	&	$-$	\\
ResNet50	&	$76.71$	&	$-0.05$	&	$-0.59$	&	$-2.17$	&	Stacked UNet-64	&	$69.53$	&	$-1.41$	&	$-3.79$	&	$-8.19$	\\
ResNet101	&	$77.50$	&	$-0.25$	&	$-0.61$	&	$-1.52$	&	SSD-ResNet18	&	$19.15$	&	$-0.81$	&	$-2.49$	&	$-5.55$	\\
ResNeXt50	&	$77.68$	&	$+0.02$	&	$-0.49$	&	$-2.01$	&	SSD-ResNet50	&	$24.93$	&	$-0.52$	&	$-2.06$	&	$-5.31$	\\
ResNeXt101	&	$79.27$	&	$+0.20$	&	$+0.19$	&	$-$	&	Faster R-CNN	&	$37.71$	&	$-0.25$	&	$-1.61$	&	$-5.51$	\\
WideResNet50	&	$78.13$	&	$+0.13$	&	$-0.34$	&	$-1.06$	&	Mask R-CNN	&	$38.26$	&	$-0.34$	&	$-1.42$	&	$-4.80$	\\
WideResNet101	&	$78.63$	&	$-0.12$	&	$+0.04$	&	$-1.00$	&	Mask R-CNN	&	$35.03$	&	$-0.88$	&	$-2.65$	&	$-7.44$	\\
InceptionV3	&	$77.10$	&	$-0.08$	&	$-0.88$	&	$-3.25$	&	Mask R-CNN 3$\times$	&	$40.78$	&	$-0.12$	&	$-1.12$	&	$-3.51$	\\
Xception	&	$79.28$	&	$+0.04$	&	$-0.28$	&	$-1.26$	&	Mask R-CNN 3$\times$	&	$37.05$	&	$-0.03$	&	$-0.81$	&	$-2.98$	\\
DenseNet121	&	$75.46$	&	$-0.46$	&	$-1.75$	&	$-4.35$	&	RetinaNet	&	$36.48$	&	$-0.42$	&	$-2.42$	&	$-6.00$	\\
DenseNet161	&	$78.77$	&	$+0.01$	&	$-0.86$	&	$-2.35$	&	RPN	&	$57.61$	&	$-0.16$	&	$-0.90$	&	$-2.56$	\\
DenseNet169	&	$76.97$	&	$+0.04$	&	$-0.96$	&	$-3.23$	&	DETR	&	$39.90$	&	$+0.10$	&	$-0.60$	&	$-0.80$	\\
VGG11-BN	&	$70.70$	&	$-0.33$	&	$-0.79$	&	$-2.24$	&	Pix2PixHD$^*$	&	$68.83$	&	$-2.27$	&	$+3.02$	&	$+3.52$	\\
VGG16-BN	&	$74.00$	&	$-0.25$	&	$-0.46$	&	$-1.75$	&	Few-Shot Vid2Vid$^*$	&	$25.78$	&	$-0.52$	&	$+0.63$	&	$+4.52$	\\
VGG19-BN	&	$74.88$	&	$+0.09$	&	$-0.48$	&	$-1.52$	&	FAZE$^*$	&	$2.94$	&	$+0.04$	&	$-0.02$	&	$+0.04$	\\
DRN-C-26	&	$75.22$	&	$-0.30$	&	$-0.74$	&	$-2.35$	&	Vaswani Base	&	$26.87$	&	$-0.68$	&	$-1.92$	&	$-3.62$	\\
DRN-C-42	&	$76.78$	&	$-0.10$	&	$-0.62$	&	$-1.98$	&	Vaswani Large	&	$28.43$	&	$-0.09$	&	$-0.92$	&	$-2.12$	\\
DRN-A-50	&	$78.30$	&	$-0.23$	&	$-0.74$	&	$-2.28$	&	Levenshtein$^*$	&	$6.16$	&	$+0.11$	&	$+0.23$	&	$+0.45$	\\
DeiT Tiny	&	$72.70$	&	$-2.81$	&	$-8.09$	&	$-16.49$	&	GNMT	&	$24.81$	&	$-0.12$	&	$+0.26$	&	$-0.15$	\\
DeiT Small	&	$80.08$	&	$-1.53$	&	$-3.75$	&	$-8.30$	&	XL Base$^*$	&	$22.88$	&	$+0.49$	&	$+2.04$	&	$+5.41$	\\
DeiT Base	&	$81.95$	&	$-0.75$	&	$-$	&	$-$	&	XL Large$^*$	&	$17.90$	&	$+0.16$	&	$+1.01$	&	$+2.65$	\\
ShuffleNetV2	&	$68.44$	&	$-0.43$	&	$-1.44$	&	$-$	&	BERT Base	&	$87.66$	&	$-0.04$	&	$-$	&	$-$	\\
MNASNet V1	&	$71.80$	&	$-1.09$	&	$-3.36$	&	$-$	&	BERT Large	&	$90.92$	&	$-0.02$	&	$-$	&	$-$	\\
\bottomrule
\multicolumn{10}{l}{$^*$ Lower means better (negative differences mean better)}
\end{tabular}}
\end{center}
\end{table}

We also draw comparisons to sparsity literature. While \textproc{Lottery}~\cite{lottery} determines at initialization which weights participate in sparse models~\cite{grasp,syncflow}, \textproc{SET}~\cite{mocanu2018} and \textproc{RigL}~\cite{mocanu2018} rewire them throughout training (see Appendix~\ref{sec:implementation}). However, we do not consider methods that move weights across layers~\cite{dettmers2019,dst} or adopt concepts similar to those explored in this paper~\cite{nmsparsity,hubara2021}.
Figure~\ref{figmethods} shows the task error (or difference in accuracy between regular and sparse models) for these, which indicates our strategy outperforms competing approaches on all tasks and model sizes by a wide margin.

\subsection{Effects of longer training}
Since conventional models can achieve higher accuracy when trained on larger data sets or with longer training schedules~\cite{roberta}, another interesting direction is to explore the effects of sparsity on models that are trained to the limits of their capacity. We train neural models longer by extending their learning rate schedules after warmup
, e.g. a schedule $1/t$ becomes $2/t$.

\begin{figure}[!t]
\begin{center}
\includegraphics[width=\linewidth]{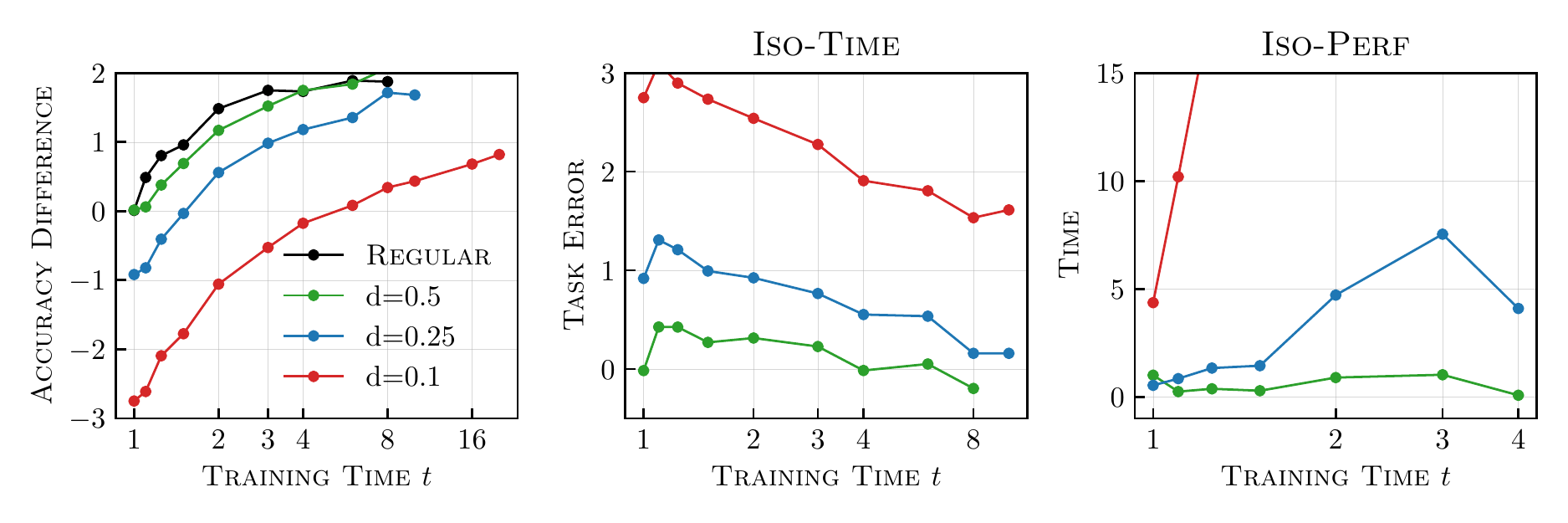}
\caption{Investigations into the effects of sparsity on longer training of ResNet18. Left: Accuracy differences as a function of training time $t$ (or ratio of steps to the original training schedule). Middle: Task error between regular and sparse models trained for the same duration. Right: Training time it takes for sparse models to match accuracy of regular models. Appendix~\ref{sec:suppscaling} covers data for a broader set of neural models and tasks.
}
\label{figlonger}
\end{center}
\end{figure}

Figure~\ref{figlonger} (left) plots accuracy differences as a function of training time $t$, which denotes the ratio of training steps to the original training schedule. The fact accuracy improves with more training suggests neural models are often not trained to capacity using conventional schedules. We observe sparse models achieve worse accuracy than regular models that are undertrained ($t\sim1$), but can match accuracy when trained to capacity ($t\gg1$).

Figure~\ref{figlonger} (middle) illustrates task errors between regular and sparse models that have been trained for the same number of steps. We find error scales inversely with $t$, since more training gives neural models better chances to explore search spaces. Particularly, errors can reach zero with sufficient training~\cite{rigl} for models that are not constrained by capacity ($d\geq0.25$). Figure~\ref{figlonger} (right) shows the time it takes for sparse models to recover accuracy of regular models is relatively short when $t\sim1$, but significantly longer as accuracy saturates ($t\gg1$). This time also increases for smaller model sizes, reminiscent of efforts that seek to undertrain massive neural models for efficiency~\cite{kaplan2020,li2020}.

\begin{table}[!b]
\caption{Accuracies for regular models and their differences for 2:4 sparse models for various $t$.}
\label{tab24}
\begin{center}
\resizebox{\textwidth}{!}{%
\begin{tabular}{lcccclcccc}
\toprule
Model & $t$ & \textproc{Regular} & \textproc{2:4 1D} & \textproc{2:4 2D} & Model & $t$ & \textproc{Regular} & \textproc{2:4 1D} & \textproc{2:4 2D} \\
\midrule
ResNet18	&	$8\times$	&	$72.17$	&	$+0.00$	&	$-0.28$	&	VGG19	&	$1\times$	&	$74.88$	&	$+0.04$	&	$-0.21$	\\
ResNet34	&	$3\times$	&	$75.14$	&	$+0.06$	&	$-0.27$	&	Xception	&	$1\times$	&	$79.28$	&	$+0.04$	&	$-0.11$	\\
ResNet50	&	$4\times$	&	$77.67$	&	$+0.05$	&	$+0.09$	&	DETR	&	$1\times$	&	$39.90$	&	$-0.30$	&	$-0.40$	\\
ResNet101	&	$3\times$	&	$78.94$	&	$-0.09$	&	$-0.31$	&	Pix2PixHD$^*$	&	$1\times$	&	$68.83$	&	$-1.34$	&	$+0.68$	\\
InceptionV3	&	$3\times$	&	$78.11$	&	$-0.11$	&	$-0.18$	&	Few-Shot Vid2Vid$^*$	&	$1\times$	&	$26.06$	&	$-0.49$	&	$+0.04$	\\
ResNext50	&	$3\times$	&	$78.36$	&	$-0.21$	&	$-0.19$	&	FAZE$^*$	&	$1\times$	&	$2.49$	&	$-0.08$	&	$-$	\\
ResNext101	&	$1\times$	&	$79.27$	&	$+0.28$	&	$+0.36$	&	GNMT	&	$1\times$	&	$24.81$	&	$+0.15$	&	$+0.09$	\\
WideResNet50	&	$1\times$	&	$78.13$	&	$-0.07$	&	$-0.08$	&	Transformer Large	&	$1\times$	&	$28.43$	&	$-0.10$	&	$-0.39$	\\
WideResNet101	&	$1\times$	&	$78.63$	&	$+0.08$	&	$-0.07$	&	BERT Large	&	$1\times$	&	$90.92$	&	$+0.03$	&	$-0.55$	\\
DRN C 26	&	$3\times$	&	$77.66$	&	$-0.05$	&	$-0.11$	&		&		&		&		&		\\
\bottomrule
\multicolumn{10}{l}{$^*$ Lower means better (negative differences mean better)}
\end{tabular}}
\end{center}
\end{table}

\subsection{Application on hardware accelerators}
We have shown earlier that by using search spaces we can reduce the size of neural models while maintaining accuracy. However, such models cannot be accelerated on modern hardware with current matrix-math pipelines~\cite{park2017,galekernels} without imposing particular structures (or positions of weights used during training and inference). On the other hand, neural structures targeting hardware acceleration (e.g., removing blocks~\cite{blocksparse}, channels or filters~\cite{wen2016,li2017}, layers~\cite{michael2019}) often degrade accuracy. This apparent tradeoff between accuracy and performance (or speed) has hindered their adoption.

Therefore, we explore whether search spaces can make hardware-aware structures more amenable for deep learning. As a case study, we consider Sparse Tensor Cores introduced in NVIDIA Ampere GPU architecture~\cite{mishra2021}, which have twice the math throughput of regular matrix operations. The hardware expects a \textproc{2:4} sparsity structure that takes at least two values to be zero for each group of four values. Appendix~\ref{sec:structure} illustrates examples of \textproc{2:4} sparsity for inference (\textproc{1D}) and for training (\textproc{2D}).

Table~\ref{tab24} lists accuracy differences between sparse and regular models (positive values mean sparse performs better) using various ratios $t$ for training schedules. We find structured neural models can generally retain accuracy for tasks. Using \textproc{2:4} for training (\textproc{2D}) and inference (\textproc{1D}) roughly match accuracy of regular models, while neural models adopting coarser structures such as block sparsity~\cite{narang2017,blocksparse} fail to benefit from search spaces and perform no better than smaller models (see Appendix~\ref{sec:structure}). Therefore, we can conclude \textproc{2:4} sparsity structures are particularly effective at approximating wider search spaces due to their finer granularity. This suggests possible avenues towards accelerating training~\cite{hubara2021} as well as obtaining efficient models for inference~\cite{mishra2021, nmsparsity}.

\section{Related work}\label{sec:relatedwork}
While this paper presents new perspectives on the benefits of using more weights, overparameterization has been extensively explored for neural model training, ranging from theoretical studies of how sufficiently large yet contrived models can converge to global minima~\cite{allenzhu2019,zou2019} to empirical investigations using more realistic examples of how training explores loss landscapes~\cite{goodfellow2015,ardalani2019}, albeit from different angles. More recently, dramatic advances have been made by training larger neural models~\cite{brown2020,fedus2021}. It was further observed that accuracy scales as a function of model size (dubbed as scaling laws~\cite{hestness2017,kaplan2020,henighan2020}) and larger models can converge to better accuracy in fewer training steps~\cite{ardalani2019,li2020}. 

Applying sparsity to reduce the size of neural models has been a topic of interest for the past three decades~\cite{lecun1990,hassibi1992,reed1993,castellano1997,han2015,topkast}. Typically, sparse models are constructed by training much larger models (that are easier to train) and removing some of their weights either after training~\cite{han2015,mishra2021} or gradually alongside training~\cite{narang2017,zhu2018,gale2019}. However, the above are only useful to reduce costs for inference.

For training acceleration, early works ~\cite{snip,grasp,syncflow} sought to remove weights before training with limited success~\cite{gale2019,frankle2020}. Surprisingly, it was shown sparse models can be trained when initialized the same way as a larger trained model~\cite{lottery}. After this breakthrough, many works~\cite{dettmers2019,rigl,topkast} tried to dynamically learn sparse models by rewiring their weights throughout training (though some works were published earlier~\cite{bellec2018,mocanu2018}), which improved accuracy but not enough to match that of larger models. More recent advances~\cite{neuralwirings,dst,savarese2020,nmsparsity,hubara2021} introduced gradient updates for all weights in a neural model, including ones that do not participate in training, and observed better success.

Our explanations about what role adding more weights might have on training -- they act as extra degrees of freedom which facilitate search for neural models -- complement existing observations that larger models (which have more weights) train better and faster. We use this understanding to consolidate many ideas that have been circling around 
in sparsity literature but remain not clearly understood (how frequently must weights be rewired and why). Recent works converging in similar directions~\cite{nmsparsity,hubara2021} or seeking better understanding~\cite{mocanu2021} are conducted in tandem with our research.

\section{Conclusion}\label{sec:conclusion}
We describe a simple reason more weights are needed for training than inference -- adding more weights gives neural models extra degrees of freedom to augment search spaces (with new paths for optimization) for training neural models effectively (or escaping critical points). Our experiments uncover recommendations to approximate the behavior of larger models (or wider search spaces) when training sparse models and demonstrate that they work across dozens of deep learning workloads.

We believe these results open many questions. On the practical side, it may be interesting to consider how could this strategy be adopted to accelerate real-world training and inference workloads today, reducing the environmental impact from training very large models. On the theoretical side, we would like to understand better ways to approximate wider search spaces in cases where accuracy still suffers as well as how to address potential biases induced by sparse models which may affect applications. We hope that our results will spur further research on these unconventional architectures, which challenge the default choice held by massive neural models today.

\section*{Acknowledgments}
We thank Paulius Micikevicius for fruitful discussions and for feedback on drafts of this work.

\bibliographystyle{unsrt}
\bibliography{searchspace}

\appendix
\newpage
\section{A primer on search spaces}\label{sec:suppsearchspace}
We expand our investigations on the role of search spaces in neural model training to cover more neural architectures and deep learning tasks.

\begin{figure}[!htb]
\begin{center}
\includegraphics[width=\linewidth]{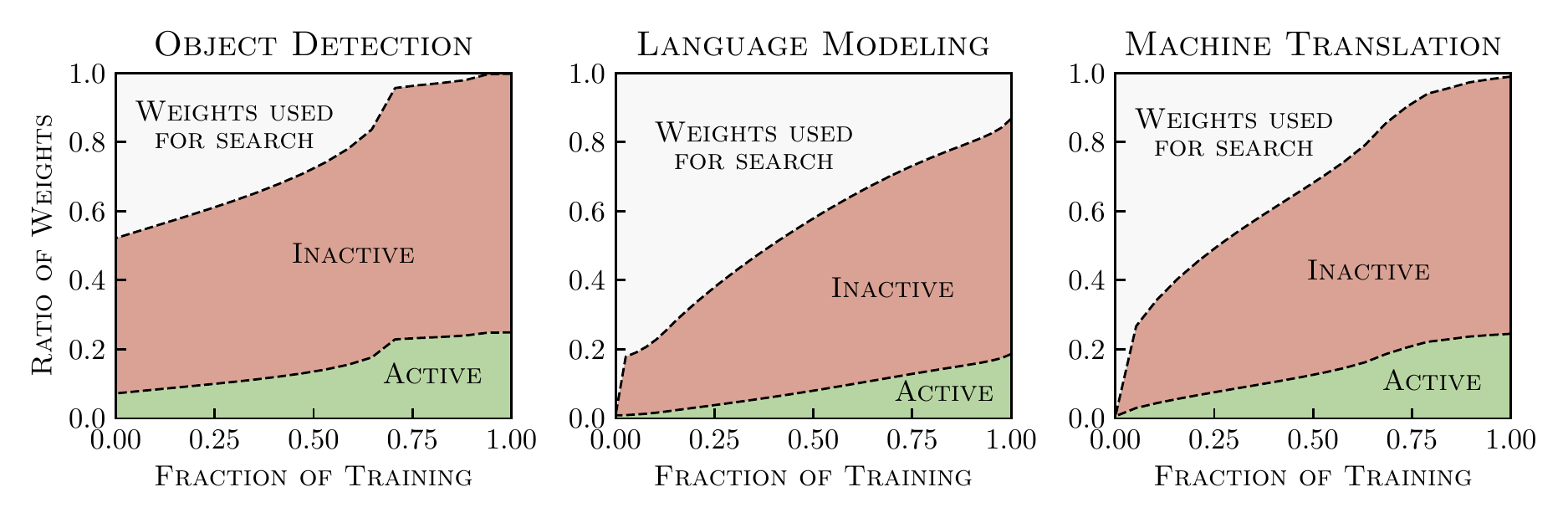}
\caption{Same as Figure~\ref{figsearch} (left). From left to right: Mask-RCNN (Object Detection), Transformer-XL (Language Modeling), GNMT (Machine Translation).}
\label{figstable}
\end{center}
\end{figure}

Figure~\ref{figstable} shows the fraction of weights whose
magnitudes remains above (active set) or below (inactive set) a threshold until the end of training, where the threshold is chosen such that one-fourth of the weights remain for inference tasks. Similar as observed in Section~\ref{sec:discovering}, we find neural weights that are needed for inference (or active set) are discovered throughout rather than early in training for other tasks as well.

\begin{figure}[H]
\begin{center}
\includegraphics[width=0.67\linewidth]{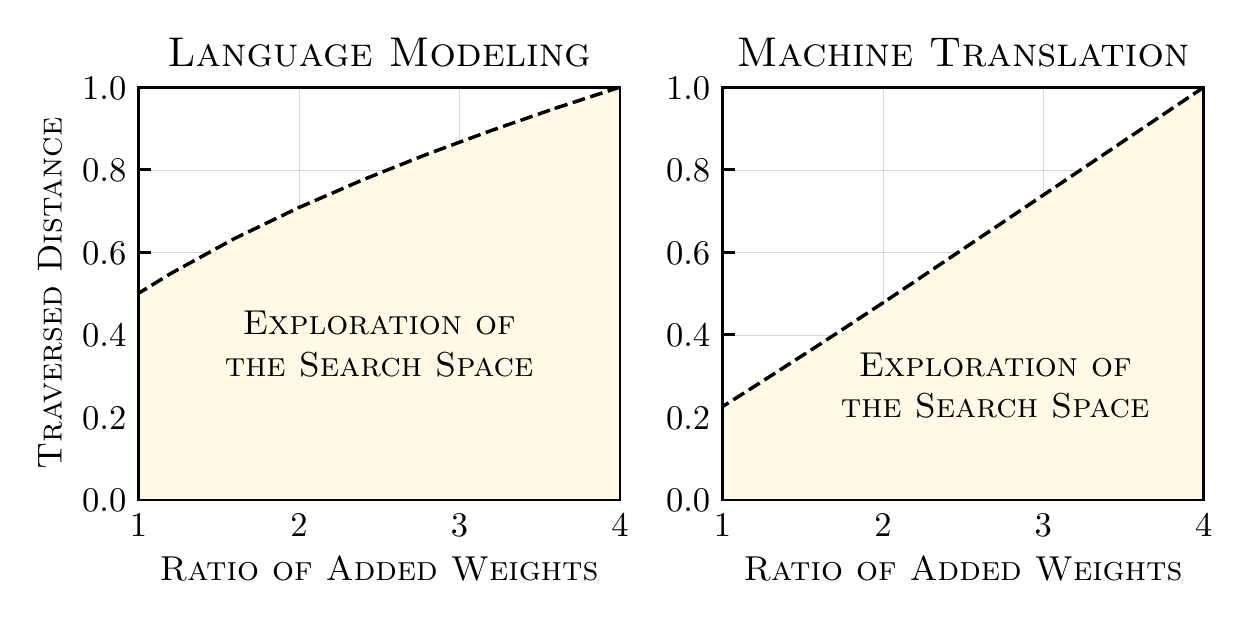}
\caption{Same as Figure~\ref{figsearch} (middle). From left to right: Transformer-XL (Language Modeling), GNMT (Machine Translation).}
\label{figdistance}
\end{center}
\end{figure}

Figure~\ref{figdistance} illustrates cumulative distances (normalized by the maximum) that neural models traverse in search space as a
function of the number of weights being added to the active set. Besides vision tasks, we find adding weights also help neural models better explore search spaces for language tasks.

\begin{figure}[H]
\begin{center}
\includegraphics[width=0.67\linewidth]{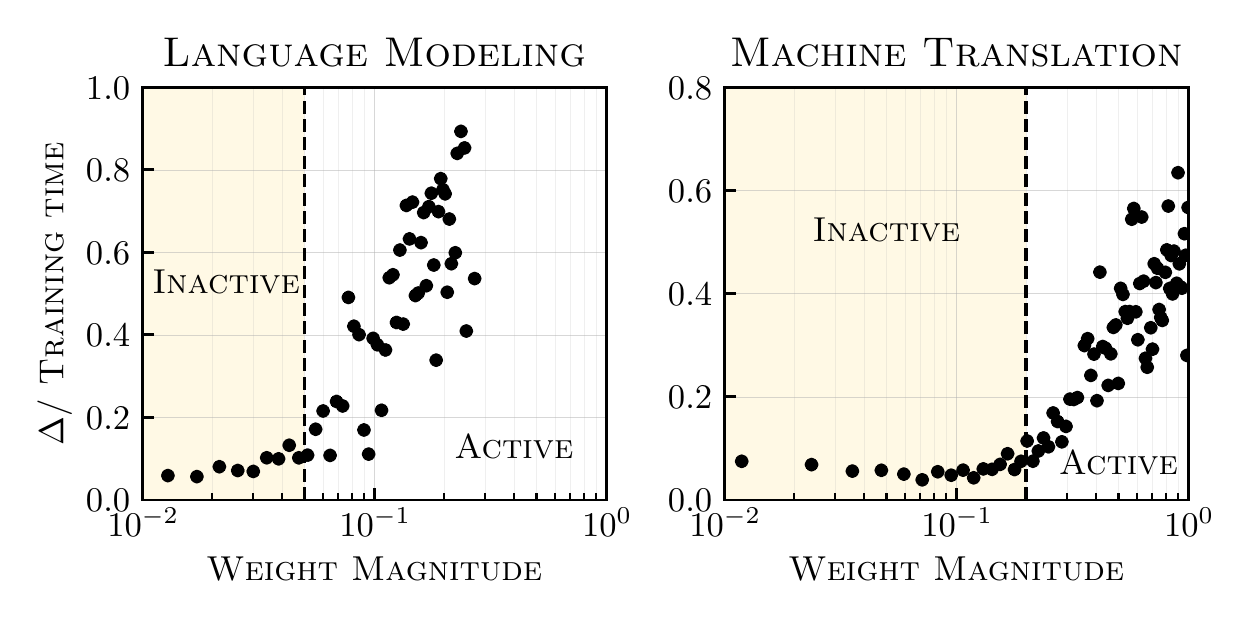}
\caption{
Same as Figure~\ref{figsearch} (right). From left to right: Transformer-XL (Language Modeling), GNMT (Machine Translation).}
\label{figcor}
\end{center}
\end{figure}

Figure~\ref{figcor} shows the fraction of time ($\Delta$) it takes for weights to decorrelate as a function of magnitudes obtained
after training. Since their correlations are short-termed, added weights (or weights that obtain small values) do not learn meaningful representations over time, rather they provide alternate paths for optimization during training to escape critical points.

Similar conclusions can be drawn as in Section~\ref{sec:searchspace}, providing more evidence that the size of search spaces influences neural model training.

\section{Recommendations for training}\label{sec:suppenhance}
We expand our investigations on how to approximate the behavior of larger neural models (or wider search spaces) when training sparse models, covering more neural architectures and deep learning tasks.

\begin{figure}[H]
\begin{center}
\includegraphics[width=\linewidth]{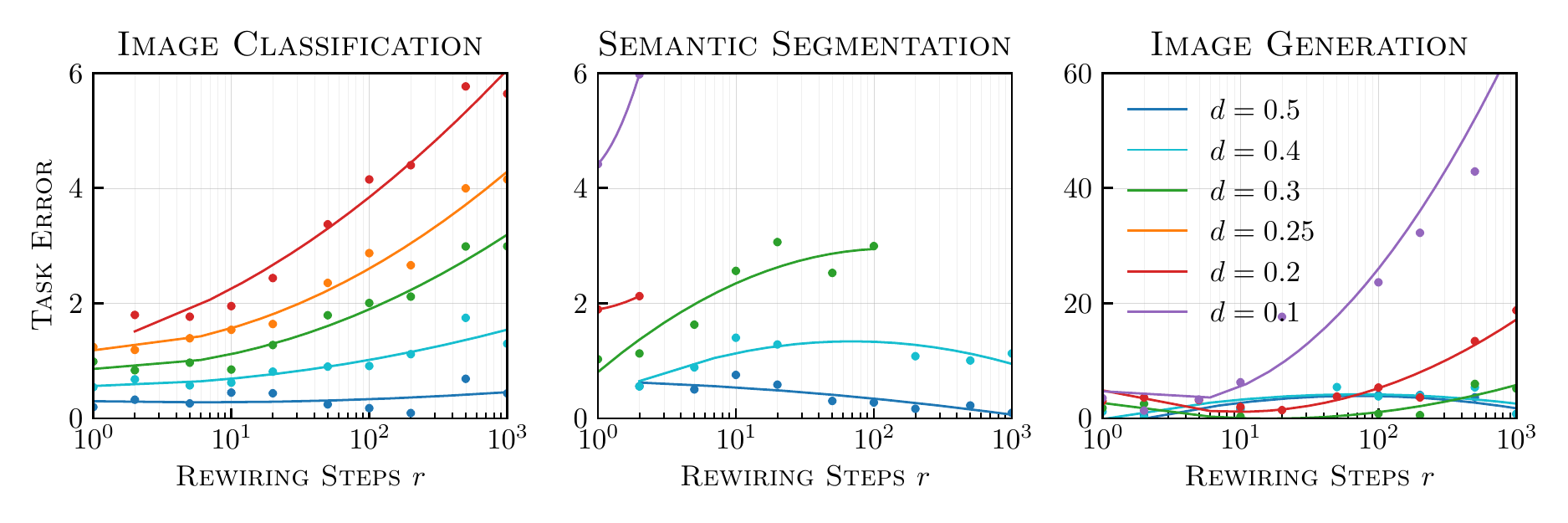}
\caption{Same as Figure~\ref{figenhance} (left). From left to right: InceptionV3 (Image Classification), Mask RCNN (Semantic Segmentation), Pix2PixHD (Image Generation).}
\label{figupdates}
\end{center}
\end{figure}

Figure~\ref{figupdates} plots the task error as a function of rewiring steps $r$ for sparse models of different sizes $d$, where $d$ denotes the ratio of weights being used. We observe that error increases with less frequent rewiring ($r\rightarrow\infty$) for other vision tasks, since rewiring is related to how often the search space expands.

\begin{figure}[H]
\begin{center}
\includegraphics[width=\linewidth]{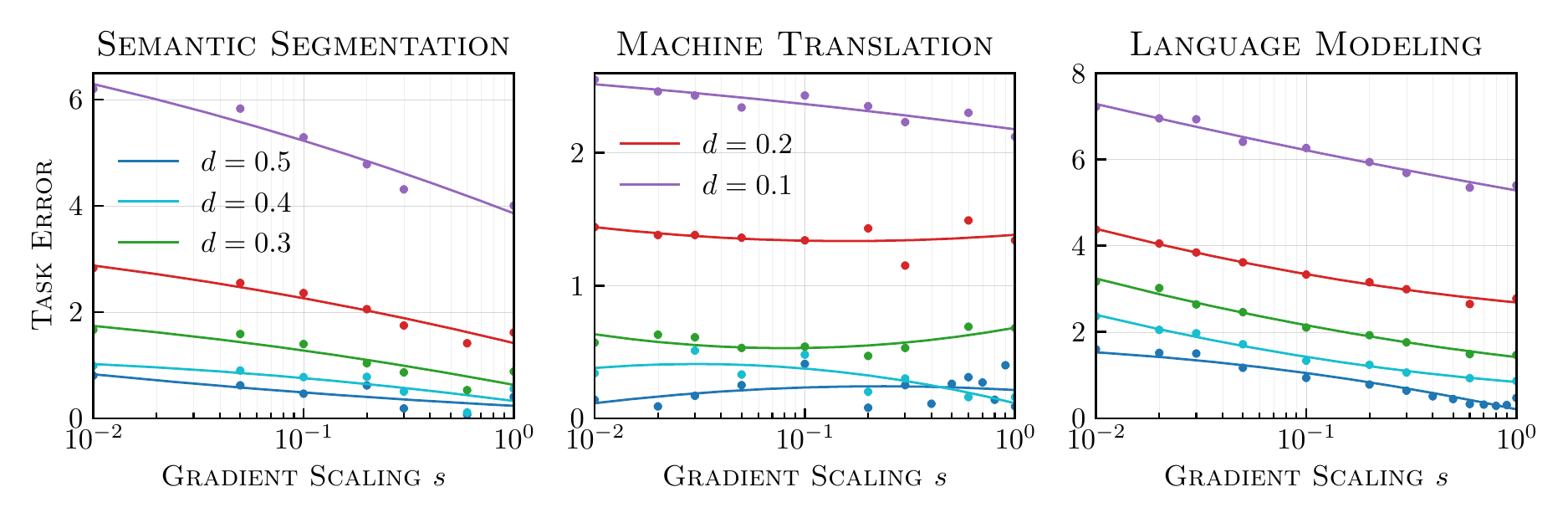}
\caption{Same as Figure~\ref{figenhance} (middle). From left to right: Mask R-CNN (Semantic Segmentation), Transformer (Machine Translation), Transformer-XL (Language Modeling).}
\label{figgrads}
\end{center}
\end{figure}

Figure~\ref{figgrads} shows the task error as a function of the scale factor $s$ applied to gradient updates for non-participating weights. We observe the error increases with decreasing contributions of the gradients ($s\rightarrow0$), which suggests updates to non-participating weights are also important during training for other tasks.

\begin{figure}[H]
\begin{center}
\includegraphics[width=\linewidth]{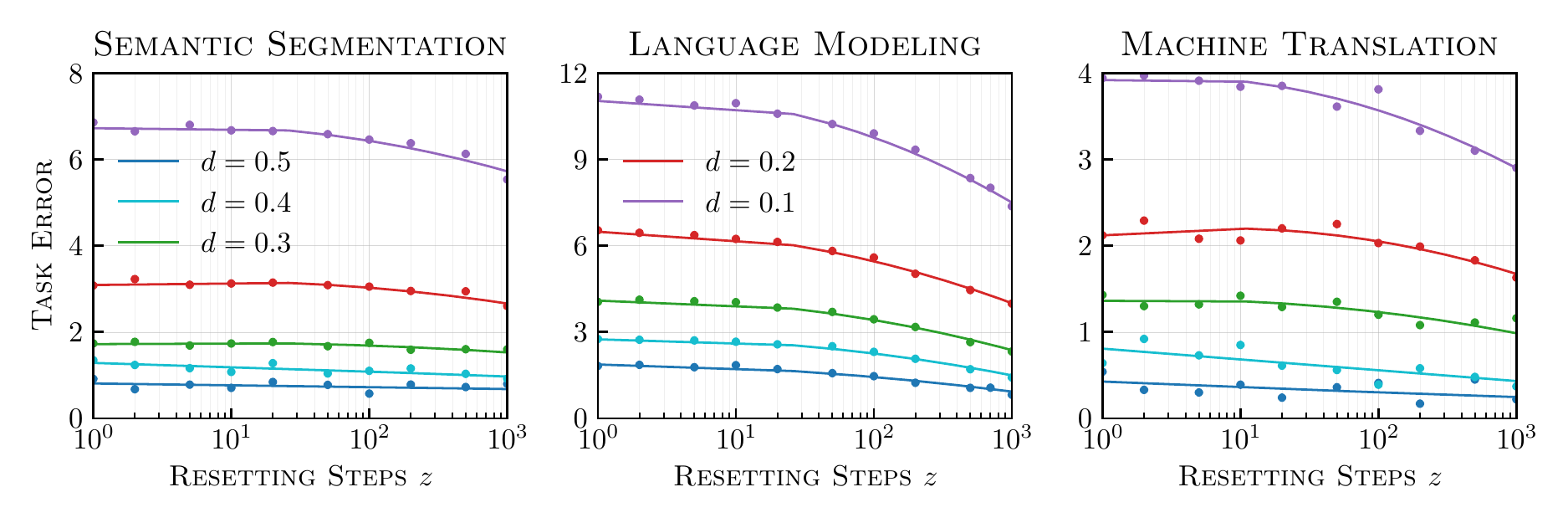}
\caption{Same as Figure~\ref{figenhance} (right). From left to right: Mask RCNN (Semantic Segmentation), Transformer-XL (Language Modeling), GNMT (Machine Translation).}
\label{figzeros}
\end{center}
\end{figure}

Figure~\ref{figzeros} demonstrates the task error as a function of the number of training steps $z$ at which non-participating weights are reset to zero. Similar to results in Section~\ref{sec:gradientupdates}, error rates saturate after sufficient training ($z\sim1k$), which reinforces the idea that non-participating weights augment search spaces rather than model capacity.

\begin{figure}[H]
\begin{center}
\includegraphics[width=\linewidth]{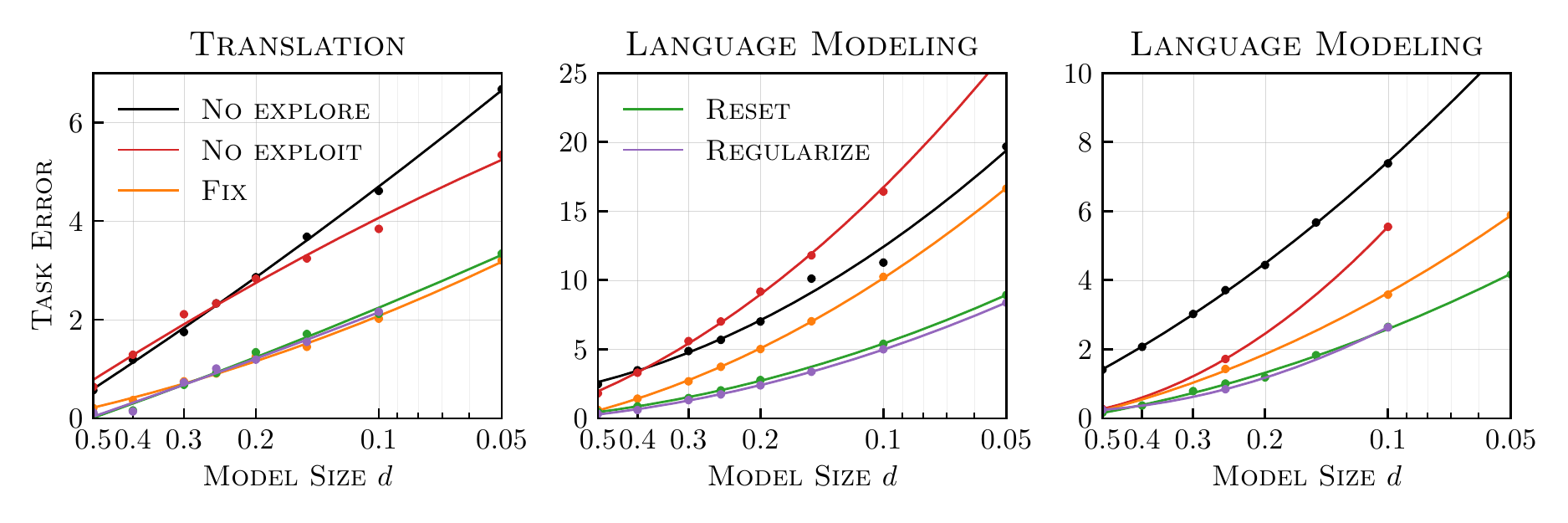}
\caption{Same as Figure~\ref{figexplorationresnet}. From left to right: Transformer (Translation), Transformer-XL Base (Language Modeling), Transformer-XL Large (Language Modeling).}
\label{figexploration}
\end{center}
\end{figure}

Figure~\ref{figexploration} compares various exploration and exploitation strategies for training sparse models, as described in Section~\ref{sec:exploitation}. While task accuracy degrades with lack of proper exploration or exploitation, inducing exploitation by removing gradient noise from non-participating weights (\textproc{Fix}, \textproc{Reset}, \textproc{Regularize}) substantially decreases the error rates.

The above results across more deep learning workloads further validate recommendations put forth in Section~\ref{sec:considerations} for training sparse models.

\section{Methodology}\label{sec:methods}
Algorithm~\ref{algo} summarizes possible methods using recommendations described in the paper.
\begin{algorithm}
\caption{\textproc{Train}($\mathcal{D},\gamma,d,v,z,\beta$)}\label{algo}
\begin{algorithmic}[1]
\State Initialize neural weights $w$ at random
\For{each training iteration $t$}
   \State Sample a mini batch of data $\mathcal{D}$
   \State $p\gets w_i$ if $w_i\geq\tau$, where $\tau$ is chosen so that $|p|=d$ \Comment{Get participating weights}
   \State $n\gets w_i$ if $w_i<\tau$ \Comment{Get non-participating weights}
   \State $\ell\gets L(p,\mathcal{D})$\Comment{Forward pass}
   \State $\frac{\partial\ell}{\partial w}\gets\frac{\partial L(p,\mathcal{D})}{\partial w}$\Comment{Backward pass}
   \State $w\gets w+\gamma\frac{\partial\ell}{\partial w}$\Comment{Optimizer step}
    \If {$t \geq v$} $n\gets 0$\Comment{\textproc{Fix}}\EndIf
    \If {$t \bmod z=0$} $n\gets 0$\Comment{\textproc{Reset}}\EndIf
    \State $n\gets n-\beta n$\Comment{\textproc{Regularize}}
\EndFor
\end{algorithmic}
\end{algorithm}

We rewire weights based on magnitude in order to preserve long-term correlations that represent learning. At each point in time, the top $d$-proportion of weights in each layer of a neural model participate in training, and the rest do not participate. We use the participating weights to compute loss and gradients, while optimizers perform gradient updates for both participating and non-participating weights (equivalent to a straight-through-estimator).

For example, the forward stage computes $x_{\ell+1}=p_\ell x_\ell$ recursively for an input $x$ and participating weights $p$, which is then fed into the loss function $L$. The backward stage derives two sets of gradients: activation gradients from layer $\ell$ are passed to downstream layer $\ell-1$ using $\partial L/\partial x_{\ell-1}=\partial L/\partial x\times p_\ell^T$, and weight gradients from layer $\ell$ are computed as $\partial L/\partial w_\ell=\partial L/\partial x_\ell\times x^T_{\ell-1}$, where $w$ denotes all of the weights in a neural model.

Training observes no memory storage savings, since we perform gradient updates for all the weights. The equations above can also only accelerate forward stages, and parts of backward stages that compute activation gradients. For further acceleration, we can consider applying sparsity to parts of backward stages that compute weight gradients~\cite{swat}.

We also note the methods above makes a tradeoff between exploration and exploitation. While approximating wider search spaces enhances more exploration, restricting non-participating weights enables exploitation. $v$, $z$, and $\beta$ dictate how and when to switch between these two phases. These variables can either be automated or kept as default, and used either in isolation or combined.

\section{Implementation details}\label{sec:implementation}
We design experiments on PyTorch~\cite{pytorch} using custom autograd functions for convolutions and linear layers. Our functions emulate sparsity using a binary tensor (or mask) that we multiply elementwise with the weights of each layer during forward and backward propagation. We determine masks based on weight magnitudes, as described earlier, after the optimizer step and before the next training iteration.

\textbf{\textproc{Search}.} We approximate wider search spaces as described in the previous appendix. We find the best tradeoff between exploration and exploitation using variables $v$, $z$, and $\beta$. Stopping rewiring after half of training ($v=0.5$) often provides the best results. $z$ should be large enough to keep short-term correlations (on the order of $1k$ training steps). $\beta=0.0002$ chosen in~\cite{nmsparsity} roughly matches our choice for $z$. All methods perform equally well with their optimal variables, so we can take different choices across workloads. For most experiments we adopt \textproc{Reset} and use \textproc{Fix} in a few select cases.

\textbf{\textproc{Reduce}.} For smaller models, we reduce the widths of neural layers by a factor of $1/d$ to match the number of weights used in sparse models. In some cases, we approximate this behavior by applying sparsity once at initialization, which represents an upper bound since tensors retain their dimensionality.

\textbf{\textproc{Lottery}.} We construct lottery tickets by training neural models to completion (e.g., $k$ steps) and computing masks based on their trained weights~\cite{lottery,lotterylarge}. We initialize sparse models with the original initialization ($t=0$) or after some amount of training ($t=\epsilon$), and train them for $k-t$ steps using the same hyperparameters. Following~\cite{lotterylarge}, we choose $\epsilon$ between $k/10$ ad $k/100$ across various workloads. While larger values for $\epsilon$ can deliver better accuracy, this comes at the cost of training acceleration, also neural models are initialized closer to their solutions~\cite{gradientflow}.

\textbf{\textproc{Set and RigL}.} We rewire participating weights over time by removing their weakest values and either adding new ones randomly~\cite{mocanu2018} or based on their gradients~\cite{rigl}. We initialize weights to zero when they become participating. The fraction $f$ of weights to rewire decays linearly throughout training, which works as well as more complex schedules (such as cosine and inverse power). We choose the best initial value for $f$ and rewiring frequency across various workloads.

\section{Experimental setup}\label{sec:setup}
\subsection{Image classification}
We train popular convolutional models like ResNets~\cite{resnet}, VGG~\cite{vgg}, Stacked U-Nets~\cite{stackedunet}, Dilated Residual Networks~\cite{drn}, Inception~\cite{inception}, MobileNet~\cite{mobilenet}, as well as vision transformers like DeiT~\cite{deit}. Training involves standard pipelines for image classification on ImageNet-2012 as described in literature and found in public code repositories. For most workloads, we adopt learning rates with linear warmups for the first 5 epochs, drop the learning rate by a factor of ten at epochs 30-60-80, and stop training after 90 epochs. A few select neural models (e.g., mobilenets and vision transformers), however, are trained for more epochs using linear or cosine schedules.

We measure model quality using top-1 classification accuracy. We apply sparsity to convolutions and linear layers with some exceptions: convolutions whose input channels are not divisible by $16$ (e.g., first convolution layer, group and depthwise separable convolutions).

\subsection{Image segmentation and detection}
Image segmentation and detection tasks include both regression and classification components. Popular detectors and segmentors are typically trained in two phases: first a backbone is trained for image classification, followed by the addition of model components that are trained for detection or segmentation. Backbones are trained on ImageNet-2012, while downstream tasks are trained on COCO. We adapt training scripts and code from Detectron2~\cite{detectron2}. 

We train neural models such as regions with convolutional neural networks (R-CNN) variants~\cite{fasterrcnn,mrcnn}, vanilla one-shot detectors~\cite{ssd}, and with focal loss~\cite{retinanet}. Convolution and linear layers encountered in pretrained backbones are sparse, like for classification tasks. Detection and segmentation heads are also targeted.

\subsection{Generative modeling}
Generative Adversarial Networks (GANs) contain two subnetworks: a generative model and a discriminative model which combine regression and discrimination tasks during training. For image and video tasks, the generator model regresses pixel colors. We explore conditional GANs for super image-to-image translation and video-to-video synthesis using Imaginaire~\cite{imaginaire}. We measure quality of generated outputs using the Frechet Inception Distance (FID). We experiment with generative neural models like Pix2PixHD~\cite{pix2pixhd}, Vid2Vid~\cite{vid2vid}, and FewShot-Vid2Vid~\cite{fewshotvid2vid}, targeting convolution and linear layers.

\subsection{Machine Translation}
We explore transformer and recurrent neural models for language translation. All models are encoder-decoder style architectures trained for English to German (En-De) translation on WMT. We adapt model and training code from Fairseq~\cite{fairseq} and NVIDIA Deep Learning Examples~\cite{deeplearningexamples}. We measure model quality using BLEU scores.

We experiment with GNMT~\cite{gnmt} and transformer-based architectures~\cite{vaswani2017,levenshtein}. All linear layers are sparse, except for embeddings and vocabulary projections.

\subsection{Language modeling}
We consider recent advances in word-level language modeling using transformer decoder (left-to-right) or encoder (bi-directional) architectures~\cite{gpt,bert}. We pretrain language models in an unsupervised fashion on WikiText-103 or Wikipedia corpus, and evaluate on downstream tasks that are zero shot or require additional finetuning. We train them using Megatron~\cite{megatron} and NVIDIA Deep Learning Examples~\cite{deeplearningexamples}. Model quality is measured in terms of perplexity or F1 score.

We train language models such as Transformer-XL~\cite{transformerxl} and BERT~\cite{bert}. We make all linear layers sparse, except for embeddings, vocabulary projections, and classification heads for downstream tasks.

\section{Search capacity}\label{sec:capacity}
It is also interesting to understand how well can our sparse models approximate wider search spaces. To this end, we measure where does accuracy of sparse models (\textproc{Search}) fall between accuracies of larger neural models with free search (\textproc{Regular}) and smaller ones that operate on reduced spaces (\textproc{Reduce}). We designate a metric $f=(\textproc{Regular}-\textproc{Search})/(\textproc{Regular}-\textproc{Reduce})$ to represents the capacity for search, where a value of one means accuracy matches that of free search, and zero implies accuracy is no better than without any additional search.

\begin{figure}[!htb]
\begin{center}
\includegraphics[width=\linewidth]{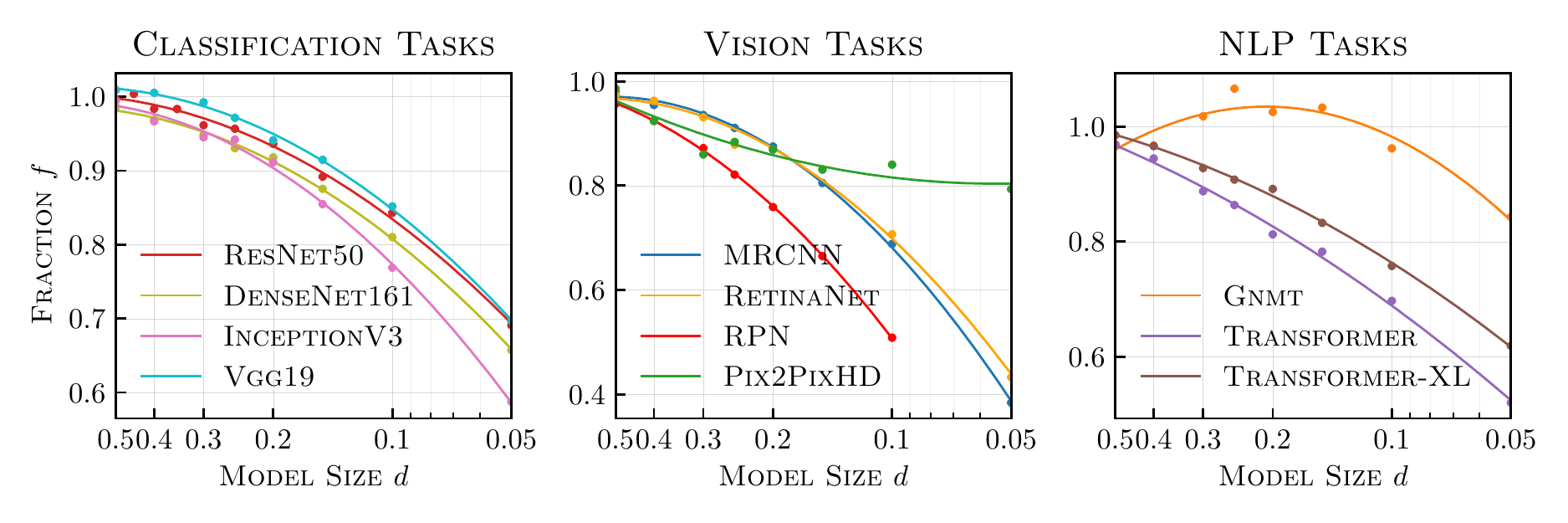}
\caption{Fraction of accuracy that sparse models achieve between regular and smaller models as a function of model size $d$ across various tasks and neural architectures. We measure this fraction as $f=(\textproc{Regular}-\textproc{Search})/(\textproc{Regular}-\textproc{Reduce})$ which represents capacity for search.
}
\label{figsparsity}
\end{center}
\end{figure}

Figure~\ref{figsparsity} illustrates $f$ as a function of size $d$ for various tasks. For $d\sim0.5$, sparse models are able to approximate wider search spaces found in larger models across all neural architectures and tasks. This helps explain why models with moderate amounts of sparsity are less sensitive to search constraints. While $f$ decreases for sparser models ($d\ll0.5$), they are still much more efficient than trivially smaller models (obtained by reducing widths of neural layers). Even at $d=0.1$, most models can capture three-quarters of the possible search capacity. 

\section{Effects of longer training}\label{sec:suppscaling}
\begin{figure}[!htb]
\begin{center}
\includegraphics[width=\linewidth]{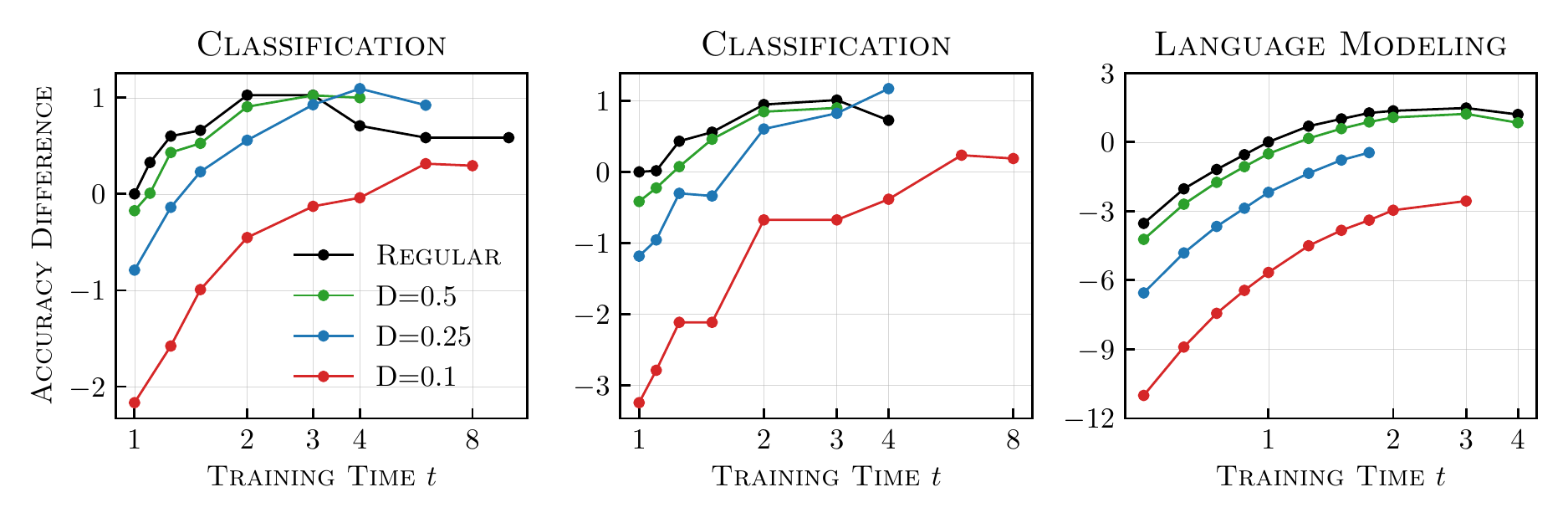}
\caption{Same as Figure~\ref{figlonger} (left). Left to right: ResNet50 (Classification), InceptionV3 (Classification), Transformer-XL (Language Modeling).}
\label{figsupplonger}
\end{center}
\end{figure}

We expand our investigations on the effects of longer training to cover more neural architectures and deep learning task. Figure~\ref{figsupplonger} illustrates accuracy deltas as a function of training time $t$. For vision tasks, we find sparse models of moderate sizes ($d\ge0.25$) can match accuracy of regular models after sufficient training $(t\ge3)$. On the other hand, for language modeling, sparse models cannot match accuracy for any time $t$ because regular models are already near capacity for the task at hand. Obviously, when using popular training schedules ($t\sim1$), sparse models can be trained a bit longer to recover the lost accuracy.

\section{Application on hardware accelerators}\label{sec:structure}
This appendix discusses various structures (or collection of weights in a sparse model) considered in the paper that are amenable for acceleration using modern matrix-math hardware.

\subsection{Block sparsity}
\begin{figure}[!htb]
\begin{center}
\includegraphics[trim=0 210 0 140,clip,width=\linewidth]{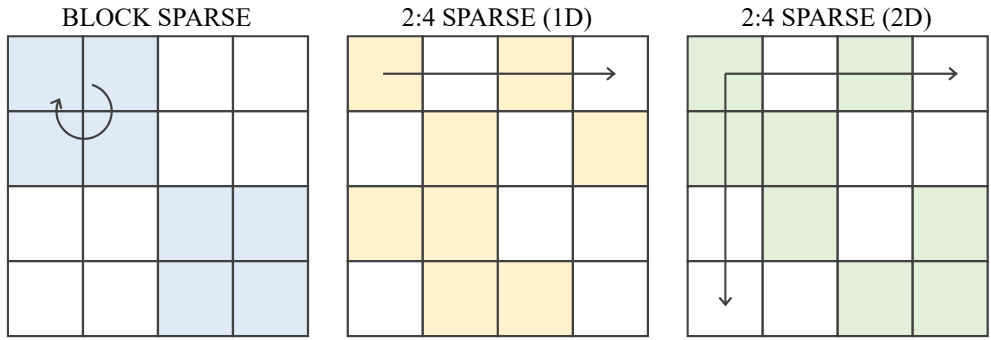}
\caption{Block sparse, \textproc{2:4 1D}, and \textproc{2:4 2D} structures for a $4\times4$ neural layer. Blank cells represent weights that do not participate in training (are assumed zero) and colored cells denote weights that participate (have nonzero values). Arrows indicate direction along which structure is imposed.}
\label{figstructure}
\end{center}
\end{figure}

We first look at block sparsity~\cite{narang2017,blocksparse}, which removes blocks of contiguous elements (or weights) in a neural layer as shown in Figure~\ref{figstructure}. Block sparsity addresses common issues that are present for unstructured formats: indices for active blocks reduce storage overhead by a factor of the block size, blocks are stored contiguously in memory which reduces irregular memory accesses, and their computations can exploit faster matrix-math hardware, such as Tensor Cores in NVIDIA GPUs.

We construct block sparse structures by (1) partitioning a neural layer into a set of blocks, (2) aggregating elements in each block into a metric, and (3) removing blocks according to some criteria based on their metrics. While we remove blocks based on largest magnitude $\max_{i\in(1,b^2)} w_i$. other choices such as the $p$-norm $(\sum_i^{b^2}|w_i|^p)^{1/p}$ achieve similar results.

\begin{table}[!htb]
\caption{Accuracies for regular models and their differences for block sparse models using different block sizes $b$.}
\label{blocksparse}
\begin{center}
\resizebox{\textwidth}{!}{%
\begin{tabular}{lcccccccc}
\toprule
Model & \textproc{Regular} & \textproc{Reduce} & $b=1$ & $b=2$ & $b=4$ & $b=8$ & $b=16$ & $b=32$ \\
\midrule
Transformer-XL	&	$22.88$	&	$-2.48$	&	$-0.82$	&	$-2.03$	&	$-2.35$	&	$-2.24$	&	$-2.36$	&	$-2.41$	\\
Transformer	&	$28.43$	&	$-0.77$	&	$-0.21$	&	$-0.75$	&	$-1.10$	&	$-1.47$	&	$-1.98$	&	$-1.94$	\\
GNMT	&	$24.81$	&	$-2.75$	&	$-0.15$	&	$-0.15$	&	$+0.02$	&	$-0.08$	&	$-0.15$	&	$+0.30$	\\
ResNet50	&	$76.71$	&	$-1.63$	&	$-0.46$	&	$-0.99$	&	$-2.19$	&	$-2.79$	&	$-$	&	$-$	\\
Mask RCNN	&	$35.03$	&	$-0.88$	&	$-0.28$	&	$-1.17$	&	$-2.43$	&	$-3.74$	&	$-4.41$	&	$-5.00$	\\
\bottomrule
\end{tabular}}
\end{center}
\end{table}

Because structures restrict the combination of weights that can be formed in a neural model, an important question is then for what block sizes $b$ (if any) can sparse models approximate wider search spaces. Table~\ref{blocksparse} lists accuracy differences between sparse and regular models for $d=0.5$ using various block sizes. We find block sparse models fail to maintain accuracy, performing no better than smaller models with neural layers of reduced widths. Notably, accuracy deteriorates for all tasks and block sizes, including smaller blocks ($b\leq4$) that are less amenable for hardware acceleration. In other words, block structures are too coarse for approximating wider search spaces. For example, different weights in a block may have different roles during training: a block that participates in training may contain weights that are not important, wheres a non-participating block may have weights that were crucial to keep. Both cases prevent sparse models from retaining weights over time that are relevant for search, and thus impacting accuracy.

\subsection{2:4 sparsity}
We next consider Sparse Tensor Cores~\cite{mishra2021} introduced in NVIDIA Ampere GPU architecture which exploit \textproc{2:4} sparsity and have twice the math throughput of regular matrix units. Figure~\ref{figstructure} shows that \textproc{2:4} sparsity mandates each group of four values must have at least two values that are zero. Typically, \textproc{2:4} is applied on weights $w$ in the forward pass, $y=wx$. However, we can also apply \textproc{2:4} on weight transposes $w^T$ for the backward pass, $\partial L/\partial x = \partial L/\partial y\times w^T$. We denote these two options as \textproc{2:4 1D} that accelerates forward pass for inference~\cite{mishra2021}, and \textproc{2:4 2D} that accelerates both forward and backward passes for training.

The \textproc{2:4} sparsity structure must always be imposed along the inner dimension of dot products. For linear layers, we apply \textproc{2:4} on a $n\times k$ weight tensor along $k$ or $n$ (for forward or backward pass, respectively). For convolutions, we apply \textproc{2:4} on a $k\times c\times r\times s$ weight tensor along input channels $c$ or $k\times r\times s$ (for forward or backward pass, respectively), where $k$ denotes output channels, $r$ and $s$ are kernel dimensions.

We can satisfy \textproc{2:4 1D} constraints by removing weights with lowest magnitudes. Since \textproc{2:4 2D} constraints have no trivial solution, we seek to minimize the cumulative magnitude of the removed weights. In other words, for each $4\times4$ block in the tensor, we construct all possible combinations of \textproc{2:4 2D} patterns, compute their $1$-norm, and choose the structure that has the largest norm.

\section{Comparisons to sparsity research}\label{sec:suppmethods}
\begin{figure}[t]
\begin{center}
\includegraphics[width=\linewidth]{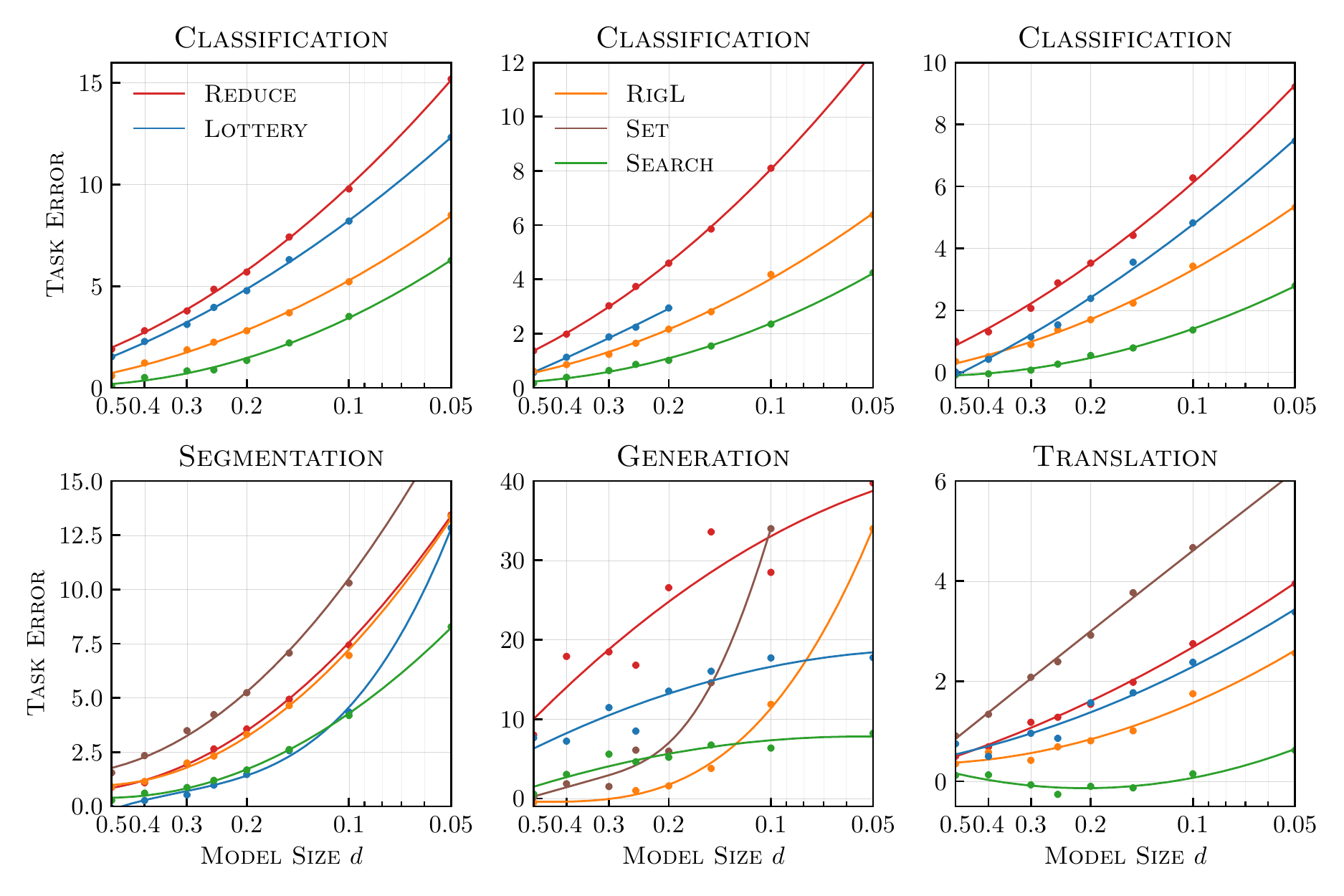}
\caption{Same as Figure~\ref{figmethods}. Clockwise: InceptionV3 (Classification), DenseNet161 (Classification), VGG19 (Classification), GNMT (Translation), Pix2PixHD (Generation), Mask RCNN (Segmentation).}
\label{figsuppmethods}
\end{center}
\end{figure}

We expand our comparisons to literature covering more neural architectures and deep learning tasks. Figure~\ref{figsuppmethods} illustrates the task error (or difference in accuracy between regular and sparse models) across various methods, neural architectures, and tasks. We find our strategy outperforms competing approaches in most cases with a few exceptions: we do not outperform lottery tickets in segmentation tasks because detectors and segmentors are trained with small learning rates which limit exploration. For generation tasks, our models are slightly worse for $d\sim0.5$ (though noisy scores make it difficult to draw conclusions), but the asymptotic behavior of error rates as $d\rightarrow0$ clearly indicates that our approach is superior.

\end{document}